\documentclass{article}

\usepackage{arxiv}

\usepackage[utf8]{inputenc} % allow utf-8 input
\usepackage[T1]{fontenc}    % use 8-bit T1 fonts
\usepackage{hyperref}       % hyperlinks
\usepackage{url}            % simple URL typesetting
\usepackage{booktabs}       % professional-quality tables
\usepackage{amsfonts}       % blackboard math symbols
\usepackage{nicefrac}       % compact symbols for 1/2, etc.
\usepackage{microtype}      % microtypography
\usepackage{graphicx}

\title{Face Detection on Surveillance Images}

\author{
  Mohammad~Iqbal~Nouyed\thanks{webpage: https://sites.google.com/site/iqbalnouyed/ ---\emph{not} for acknowledging funding agencies.} \\
  Lane Department of Computer Science and Electrical Engineering\\
  West Virginia University\\
  Morgantown, WV 26505 \\
  \texttt{monouyed@mix.wvu.edu} \\
   \And
 Guodong~Guo \\
  Lane Department of Computer Science and Electrical Engineering\\
  West Virginia University\\
  Morgantown, WV \\
  \texttt{guodong.guo@mail.wvu.edu} \\
}

\begin{document}
\maketitle

\begin{abstract}
In last few decades, a lot of progress has been made in the field of face detection. Various face detection methods have been proposed by numerous researchers working in this area. The two well-known benchmarking platform: the FDDB and WIDER face detection provide quite challenging scenarios to assess the efficacy of the detection methods. These benchmarking data sets are mostly created using images from the public network ie. the Internet. A recent, face detection and open-set recognition challenge has shown that those same face detection algorithms produce high false alarms for images taken in surveillance scenario. This shows the difficult nature of the surveillance environment. Our proposed body pose based face detection method was one of the top performers in this competition. In this paper, we perform a comparative performance analysis of some of the well known face detection methods including the few used in that competition, and, compare them to our proposed body pose based face detection method. Experiment results show that, our proposed method that leverages body information to detect faces, is the most realistic approach in terms of accuracy, false alarms and average detection time, when surveillance scenario is in consideration.  
\end{abstract}

% keywords can be removed

\section{Introduction}
Face detection algorithms in recent years have advanced enough to hold robust against a large number of unconstrained environment criteria such as pose, occlusion, focus, illumination, and size etc \cite{lu2012survey,zhang2010survey,ZAFEIRIOU2015survey,yang2002survey}. The website of most popular face detection benchamrking platform Face Detection Data Set and Benchmark (FDDB) \cite{jain2010fddb} proposed by Jain and Learned-Millder currently displays benchmark results of $39$ published methods and $31$ unpublished ones. Recently, a new face detection benchmark called WIDER face \cite{yang2016wider} was proppsed by Yang et al. with a much more challenging dataset. 

\begin{figure}[!htbp]
\centering
\includegraphics[width=\linewidth]{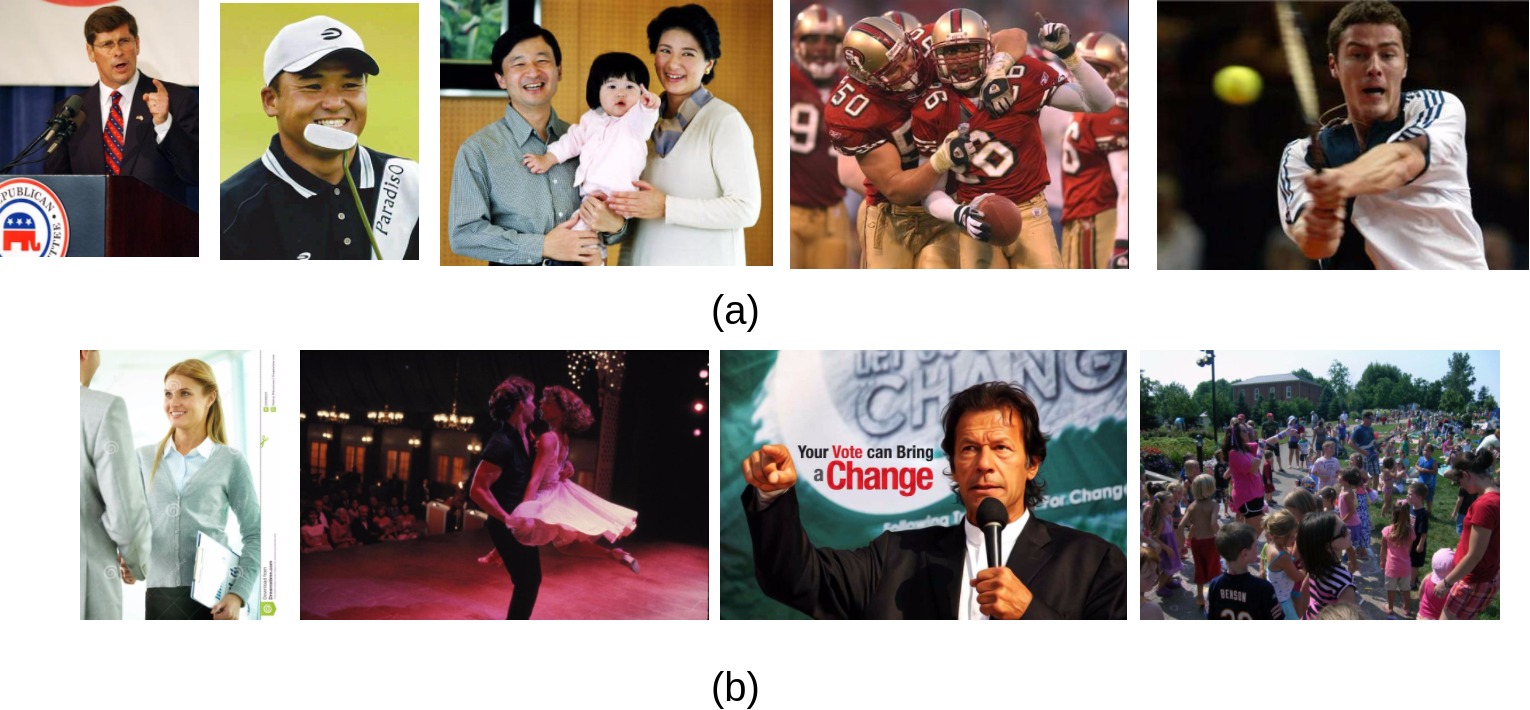}
\caption{Sample images from (a) FDDB and (b) WIDER FACE database}
\label{fig:sample_images}
\end{figure}

The FDDB database contains $2845$ images with a total of $5171$ faces. These images were collected from Yahoo! news website, and later on cleaned and annotated. The WIDER FACE dataset consists of $32,203$ images with $393,703$ labeled faces. These images were retrieved using search engines like Google and Bing. Then, they were manually cleaned by human inspection and annotated. Sample images from the both datasets are shown in Figure~\ref{fig:sample_images}.

From manual observation, we can see that eventhough these databases contain wide range of variation in pose, occlusion, lighting, face size, sharpness etc. most of them are not representative of face images collected in surveillance scenerio. Like FDDB and WIDER FACE most face detection or recognition databases the majority of images are “posed”, i.e. the subjects know they are being photographed, and/or the images are selected for publication in public media. Hence, blurry, occluded and badly illuminated images are generally uncommon in these datasets. 

This year, a new type of face detection challenge was organized which dealt with faces from surveillance scenario \cite{gunther2017uccs}. $5$ different detection algorithms were submitted in which our body pose based face detection was one of the top performing along with Tiny Faces \cite{hu2016finding}. In this paper, we tried to investigate further to understand the reason behind this high performance, which was not covered in the paper of Gunther et al. \cite{gunther2017uccs}. We perform a comparative face detection analysis with our body based based face detection method on UCCS dataset using $8$ different face detection algorithms including the other top performing algorithm Tiny Faces. Following sections provides description of the database used, face detection methods evaluated and our findings.

\begin{figure}[!htbp]
\label{fig:UCCS_images}
\centering
\includegraphics[width=.5\linewidth]{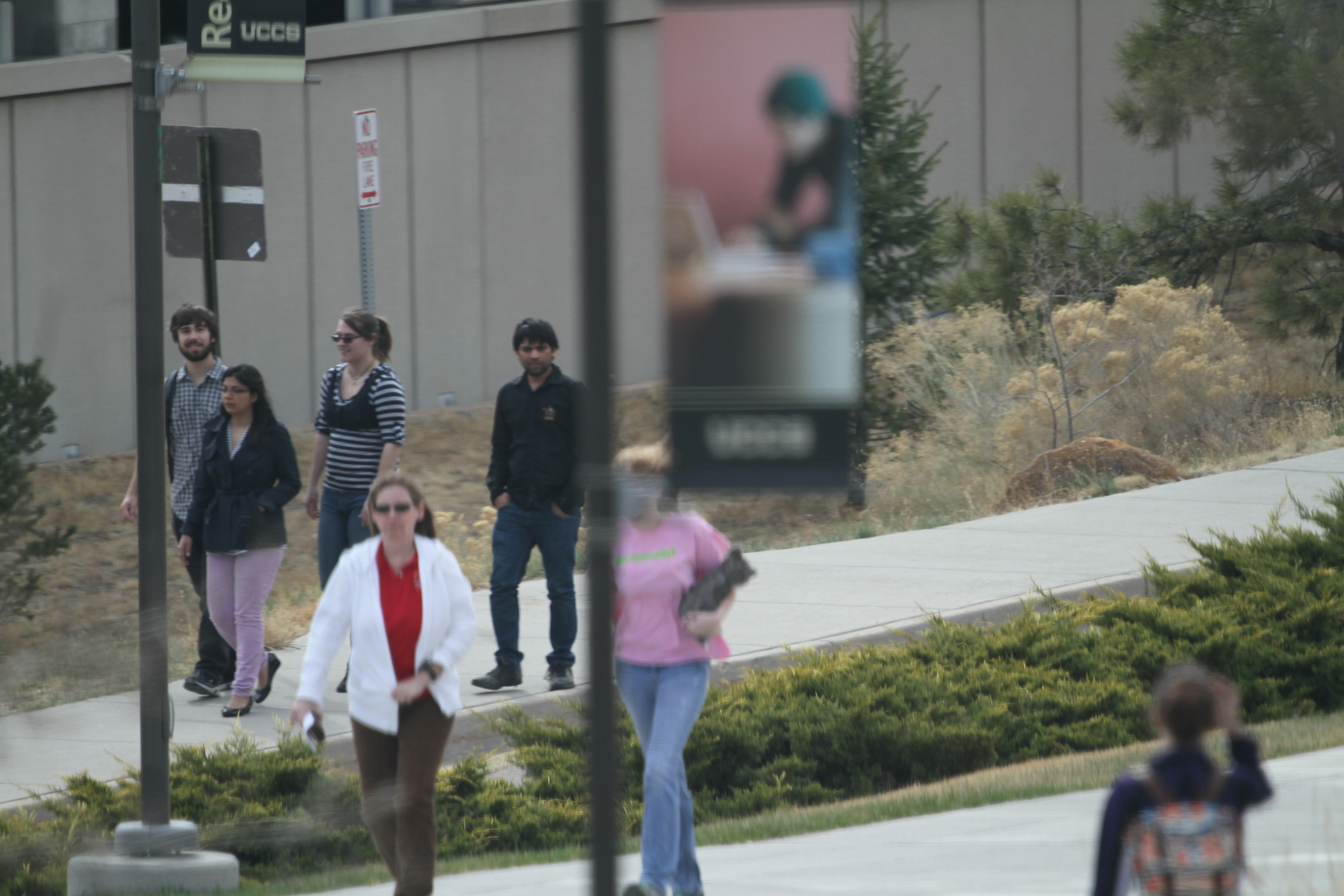} 
\includegraphics[width=.5\linewidth]{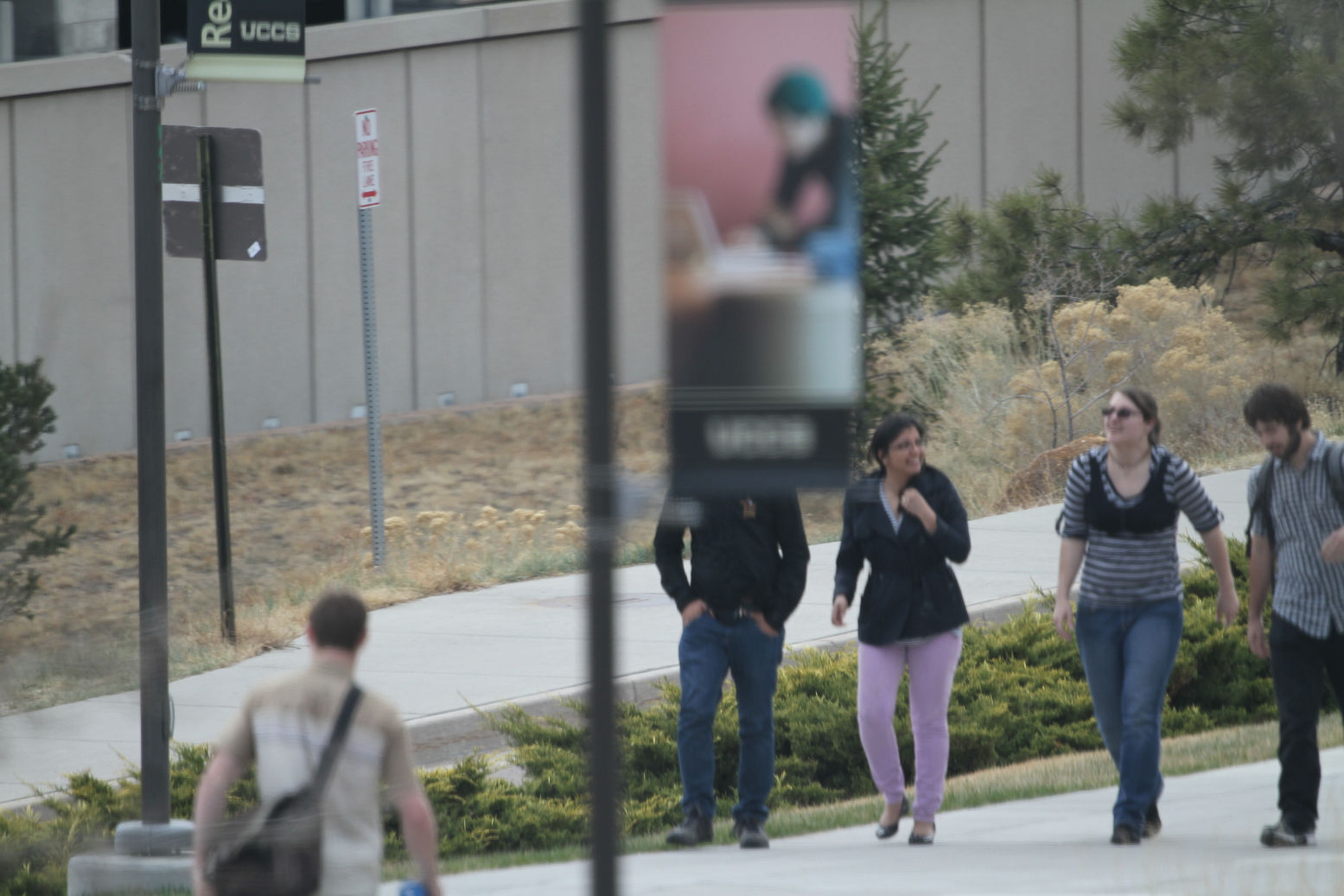}
\caption{Sample images UCCS database. Note that not a single face in these two images is frontal and without occlusion – some have small occlusion, others large; some have significant yaw and pitch angles; and many are blurred.}
\end{figure}

\begin{figure}[!htbp]
\centering
\includegraphics[width=.5\linewidth]{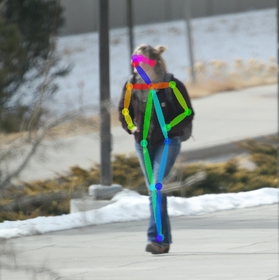}
\caption{18 Body joints detected by the body pose detector shown in colored lines.}
\label{fig:openpose}
\end{figure}

\section{UnConstrained College Students (UCCS) dataset}

In UnConstrained College Students (UCCS) dataset \cite{sapkota2013uccs}, subjects are photographed using a long-range high-resolution surveillance camera without their knowledge. Faces inside these images are of various poses, and varied levels of blurriness and occlusion. 

Images are acquired using Canon 7D camera fitted with Sigma 800mm F5.6 EX APO DG HSM lens. The camera is placed inside an office room and is focused on the outdoor sidewalk at 100m distance from the office room, resulting in 18 Megapixels scene images. Images are captured at an interval of one frame per second, during times when many students were walking on the sidewalk.. The chances of the same person appearing in front of the camera the next day at the same interval is high. For example, a student taking Monday-Wednesday classes at 12:30 PM will show up in the camera on almost every Monday and Wednesday. This results in multiple sequences of an individual on multiple days. The images contain various weather conditions such as sunny versus snowy days. They also contain various occlusions such as sunglasses, winter caps, fur jackets, etc., and occlusion due to tree branches, poles, etc.

Capturing of images was performed on $20$ different days, between February $2012$ and September $2013$ covering various weather conditions such as sunny versus snowy days. They also contain various occlusions such as sunglasses, winter caps, fur jackets, etc., and occlusion due to tree branches, poles, etc. To remove the potential bias of using automated face detection (which selects only easy faces), more than $70,000$ face regions were hand-cropped. From these, we have labeled in total $1732$ identities. Each labeled sequence contains around 10 images. For approximately $20$\% of the identities, we have sequences from two or more days. Dataset images are in JPG format with an average size of $5184\times3456$.

\section{Face Detection Methods}

In this section, we describe the face detection methods we use to compare with our proposed body based face detection method. 8 different face detection methods are compared with BFD on the UCCS dataset. 

\subsection{Tiny Faces}

This method was proposed by Hu and Ramanan \cite{hu2016finding} focuses on three aspects of the face detection problem: the role of scale invariance, image resolution, and contextual reasoning. The authors train separate detectors for different scales in a multi-task fashion. They make use of features extracted from
multiple layers of single (deep) feature hierarchy. While training detectors for small objects context is crucial. The authors define templates that make use of massively-large receptive fields.

\begin{figure*}[!htbp]
\centering
\includegraphics[width=\linewidth]{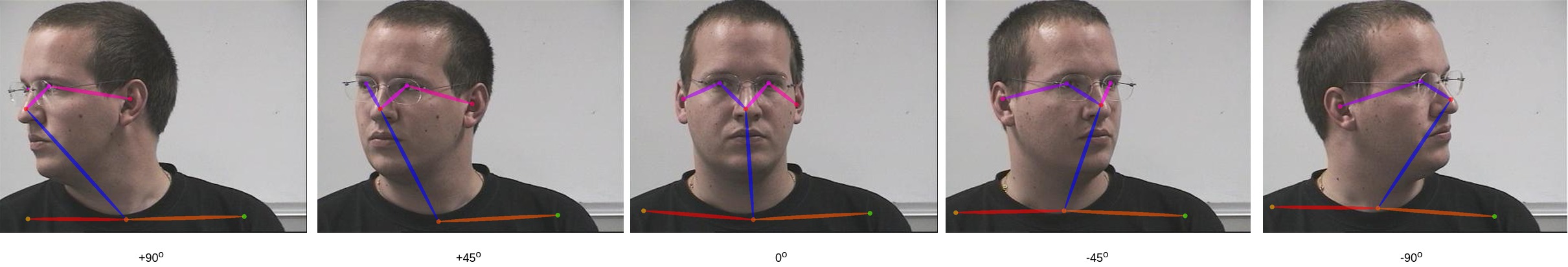}
\caption{Head pose detection is done using ear and eye joints. Here all the detected face joints are shown for different head poses. Notice, how the distance between ear and eye varies as head pose changes.}
\label{fig:face_joints}
\end{figure*}

\subsection{Multitask Cascaded Convolutional Neural Networks (MTCNN)}

Proposed by Zhang et al. \cite{zhang2016mtcnn}, the method consists of a deep cascaded multi-task framework which exploits the inherent correlation between detection and alignment to boost up their performance. This framework leverages a cascaded architecture with three stages of deep convolutional networks to predict face and landmark location in a coarse-to-fine manner. 

\subsection{Aggregate Channel Feature (ACF) based face detection }

This method was proposed by Yang et al. \cite{yang2014ffd}. They adopted the
concept of channel features to the face detection domain, which extends the image channel to diverse types like gradient magnitude and oriented gradient histograms and therefore encodes rich information in a simple form. Aggregate channel features, make a full exploration of feature design, and discover a multiscale version of features with better performance.

\subsection{Normalized Pixel Difference (NPD) based face detection}

Liao et al. \cite{liao2016npd} proposed this method. NPD feature is computed as the
difference to sum ratio between two pixel values, inspired by the Weber Fraction in experimental psychology. This feature is scale invariant, bounded, and is able to reconstruct the original image. The authors use a deep quadratic tree to learn
the optimal subset of NPD features and their combinations, so that complex face manifolds can be partitioned by the learned rules. This way, only a single soft-cascade classifier is needed to handle unconstrained face detection. 

\subsection{YOLO model based face detection}

Proposed by Redmon et al. \cite{redmon2016yolo} YOLO is a state-of-the-art, real-time object detection system. This method takes the object detection as a regression problem to spatially separated bounding boxes and associated class probabilities. A single neural network predicts bounding boxes and class probabilities directly from full images in one evaluation. Since the whole detection pipeline is a single network, it can be optimized end-to-end
directly on detection performance.YOLO learns very general representations of objects. We used a fine tuned version of the YOLO object detection model trained on a face database. 

\subsection{CNN cascade for  face detection}

Proposed by Li et al. \cite{li2015cnncascade} this method uses a cascade architecture built on convolutional neural networks (CNNs) with very powerful discriminative capability, while maintaining high performance. The CNN cascade operates at multiple resolutions, quickly rejects the background regions in the fast low resolution stages, and carefully evaluates a small number of challenging candidates in the last high resolution stage. To improve localization effectiveness, and reduce the number of candidates at later stages, a CNN-based
calibration stage is used after each of the detection stages in the cascade. 

\subsection{Face detection using Deformable Part Model (DPM)}

This method was proposed by Mathias et al. \cite{Mathias2014dpm} adopts the integral channels detector to the task of face detection. It uses simple gradient magnitude channels, and color channels. It also uses shallow boosted trees of depth two and combine a set of to represent the face components. Each component captures a fraction of the intra-class diversity of faces. A total of 5 components are used (a frontal face detector, two side views, and their mirrored versions). The component detector consists of an Adaboost classifier with 2000 weak learners. The DPM components uses those same three views, but due to mirroring has 6 components and each component has one root template and 8 parts.

\subsection{Constrained Local Neural Fields (CLNF) based face detection}

Proposed by Baltrusaitis et al. \cite{tadas2013clnf,tadas2016openface} this method uses a probabilistic patch expert (landmark detector) that can learn non-linear and spatial relationships between the input pixels and the probability of a landmark being aligned. The model is optimised using a novel Non-uniform Regularised Landmark Mean-Shift optimisation technique, which takes into account the reliabilities of each patch expert. The model detects 68 facial landmarks which are used to find out the bounding box region.

\begin{figure*}[!htbp]
\centering
\includegraphics[width=\linewidth]{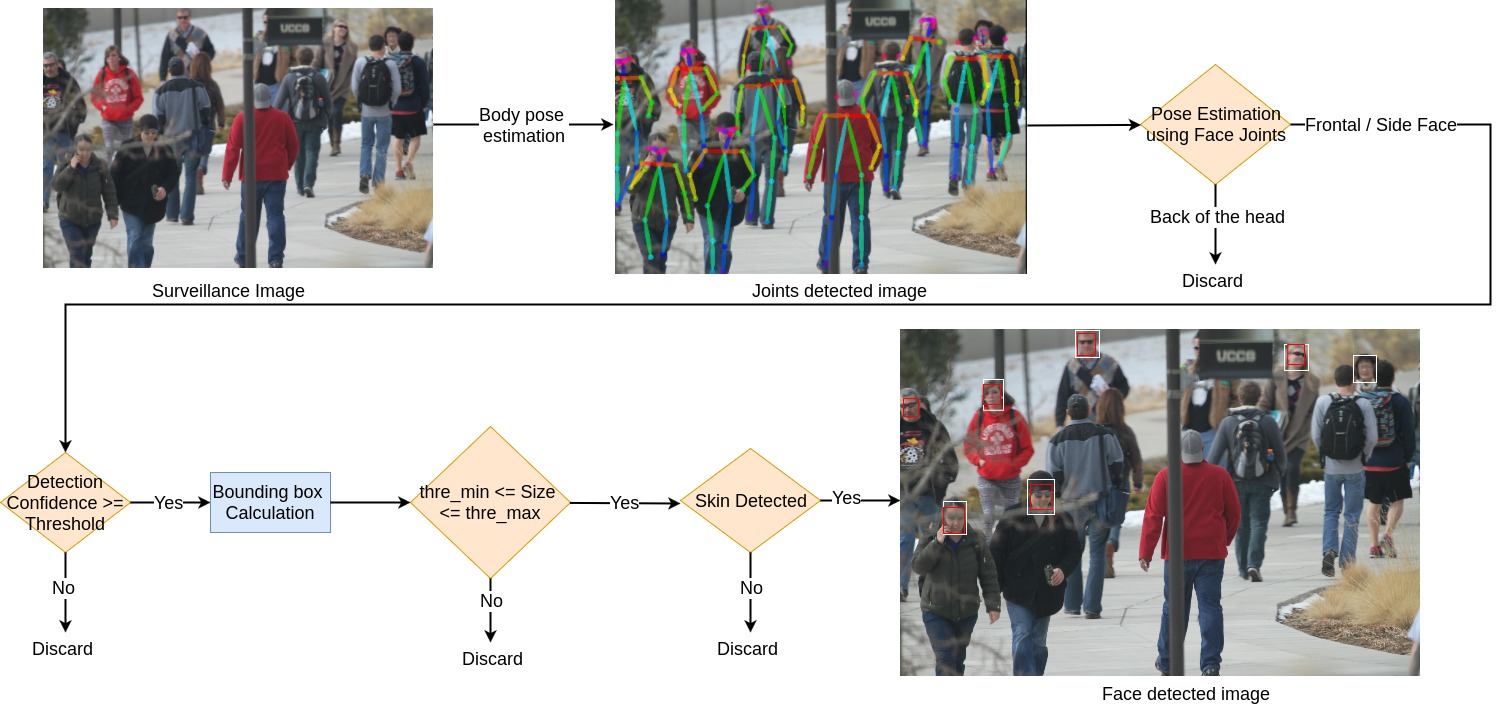}
\caption{Overview of the body based face detection method.}
\label{fig:bfd}
\end{figure*}

\subsection{Our Proposed Method: Body based face detection (BFD)}

We propose a body pose estimation based approach for real-time face detection. The approach is inspired by the human 2D pose estimation methods based on Convolutional Neural Networks (CNNs) in \cite{cao2017realtime} proposed by Cao et al. This method efficiently detects the 2D pose of multiple people in an image. it uses a non-parametric representation, which referred as Part Affinity Fields (PAFs), to learn to associate body parts with individuals in the image. The architecture encodes global context, allowing a greedy bottom-up parsing step that maintains high accuracy while achieving real-time performance, irrespective of the number of people in the image. The architecture is designed to jointly learn part locations and their association via two branches of the same sequential prediction process. The face detection algorithm is based on detected joints on face. From our experiment, this approach can be utilized for real-time face detection with high accuracy. 

The following three steps briefly summarize the processing of the real-time face detection approach: 

1) Extract coordinates of joints. Apply multi-pose estimation to the target image. The output are coordinates of 18 main joints of human body, such as shoulder center, waist and nose, etc. If any joint is undetected, the coordinates of the joint will be set to null. Figure~\ref{fig:openpose} shows the detected joints in color lines.

2) Apply frontal/side face detection based on the details of the joints and draw the boundary boxes for all detected frontal/side faces. Based on the information of the five joints on face (nose, left eye, right eye, left ear and right ear) , decide if there is a frontal/side face or back of head. We use the difference between the two eye-to-ear distances to estimate the pose. Figure~\ref{fig:face_joints} shows how these distances varies with head pose change. In order to reduce false alarm, a confidence threshold is set for face detection. The threshold is applied to all detected joints of the face, and then delete the joints whose confidence is lower than the threshold. We consider all different detection situation (angle of the face) and build a well defined frontal/side face detection rule.    

3) Apply boundary box size check to detected faces to decrease the false alarm rate. Two threshold ($thre\_min$ and $thre\_max$) are set for checking the size of boundary box of each face. If  $boundary \ box>thre\_max$ or $boundary \ box<thre\_min$, we delete such detected face. Here we set $thre\_min=90$ and $thre\_max=500$.

4) Finally, a skin detector method was trained using part of the training set. This helped to remove more false alarms by thresholding the distance between the distribution of the skin color of face image, from the distribution of the training set. 

Figure~\ref{fig:bfd} shows the flow diagram of this method.

\begin{figure*}[!htbp]
\centering
\includegraphics[width=0.4\linewidth]{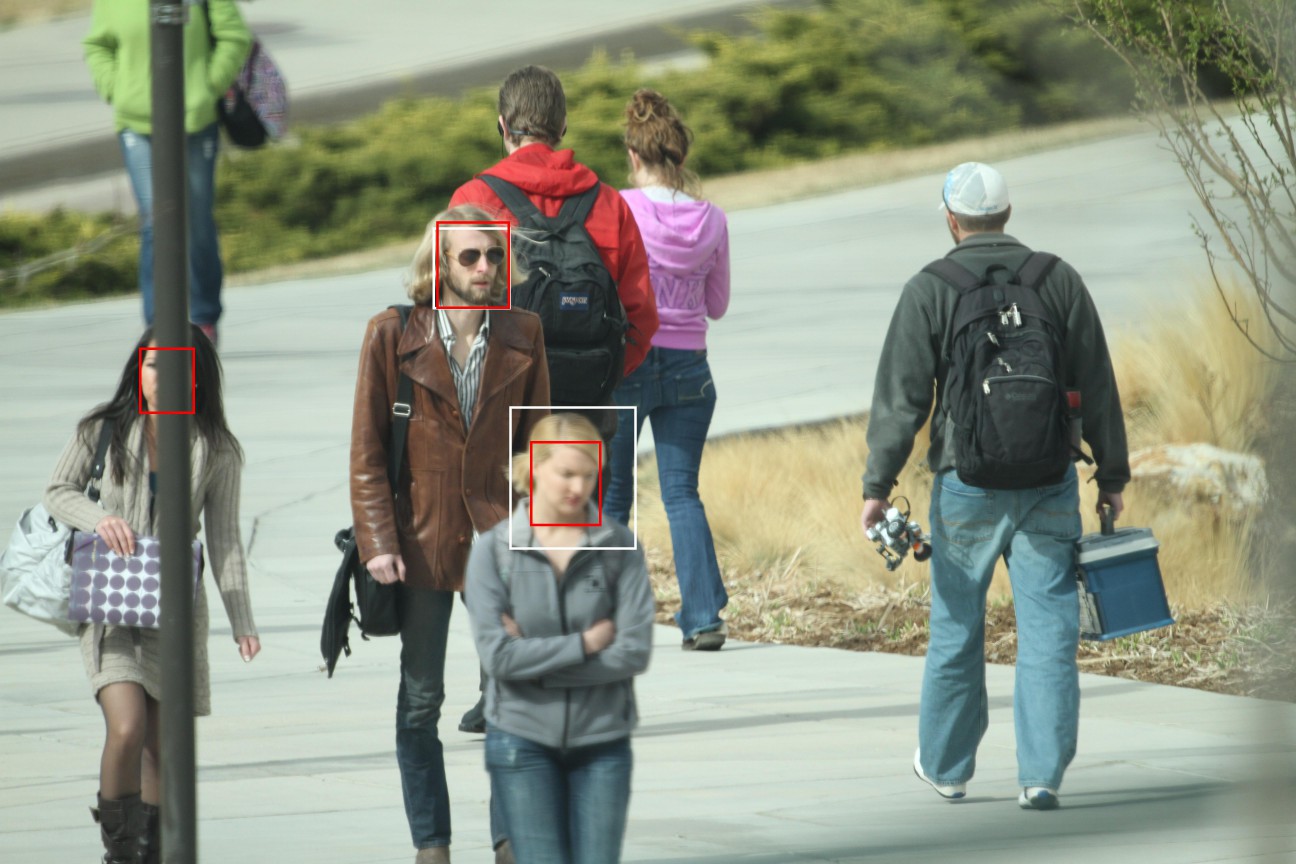}
\includegraphics[width=0.4\linewidth]{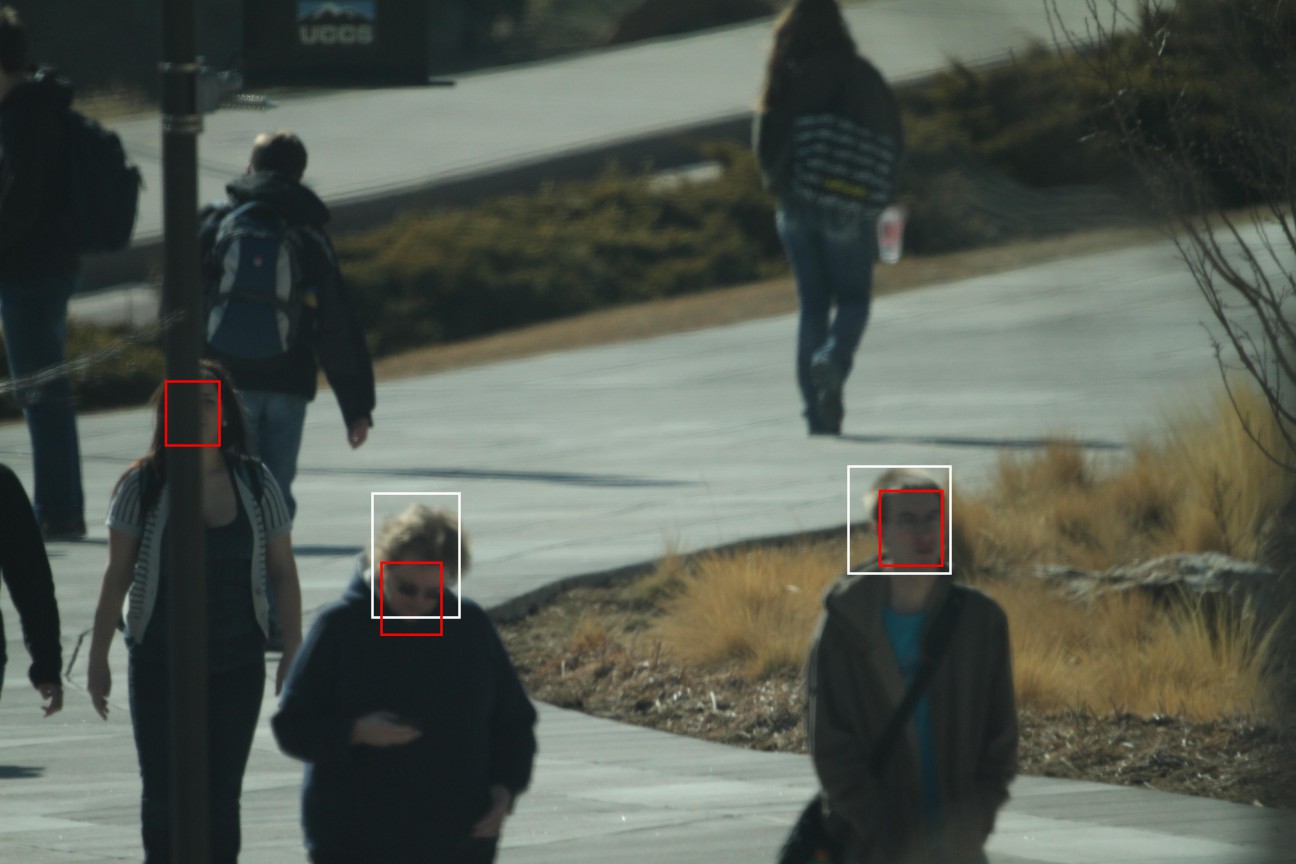} \\
\includegraphics[width=0.4\linewidth]{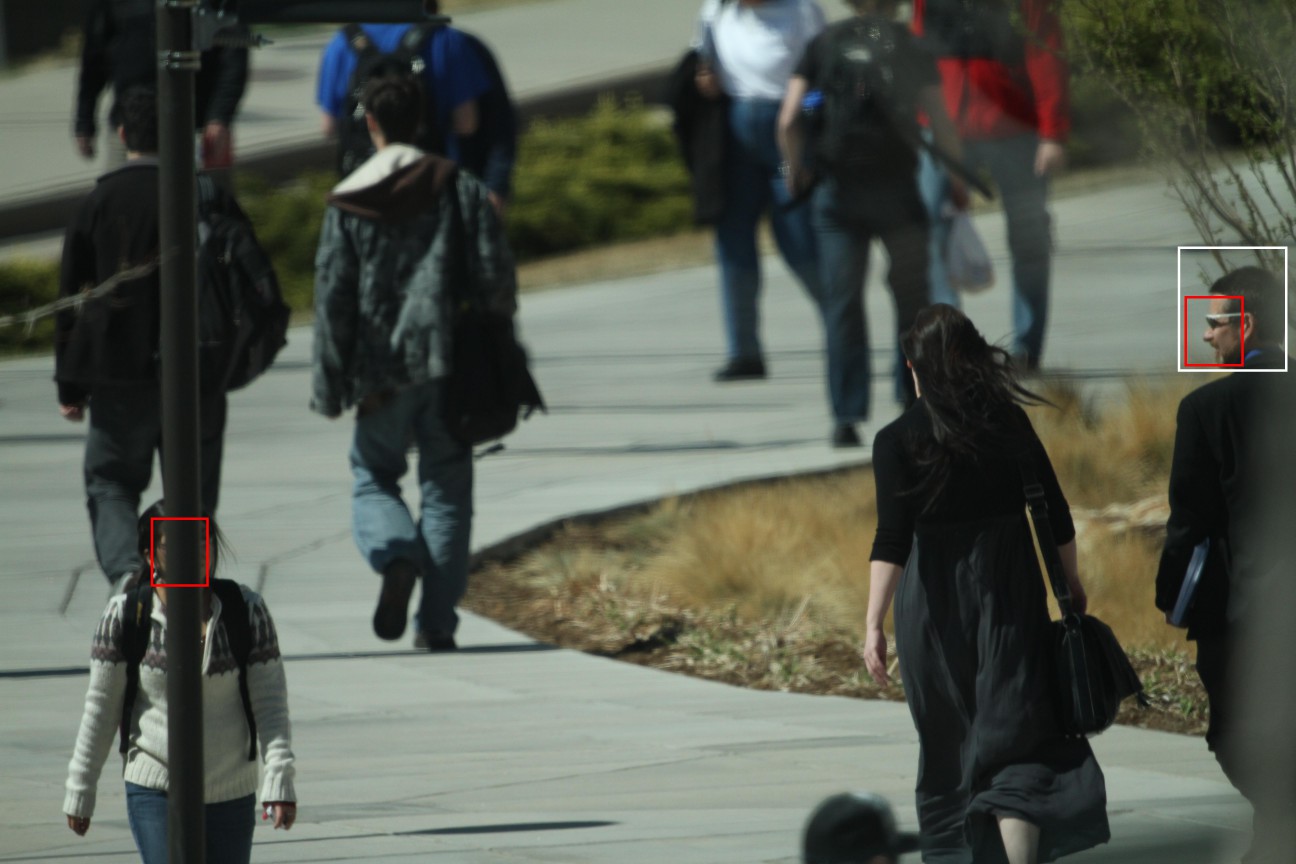}
\includegraphics[width=0.4\linewidth]{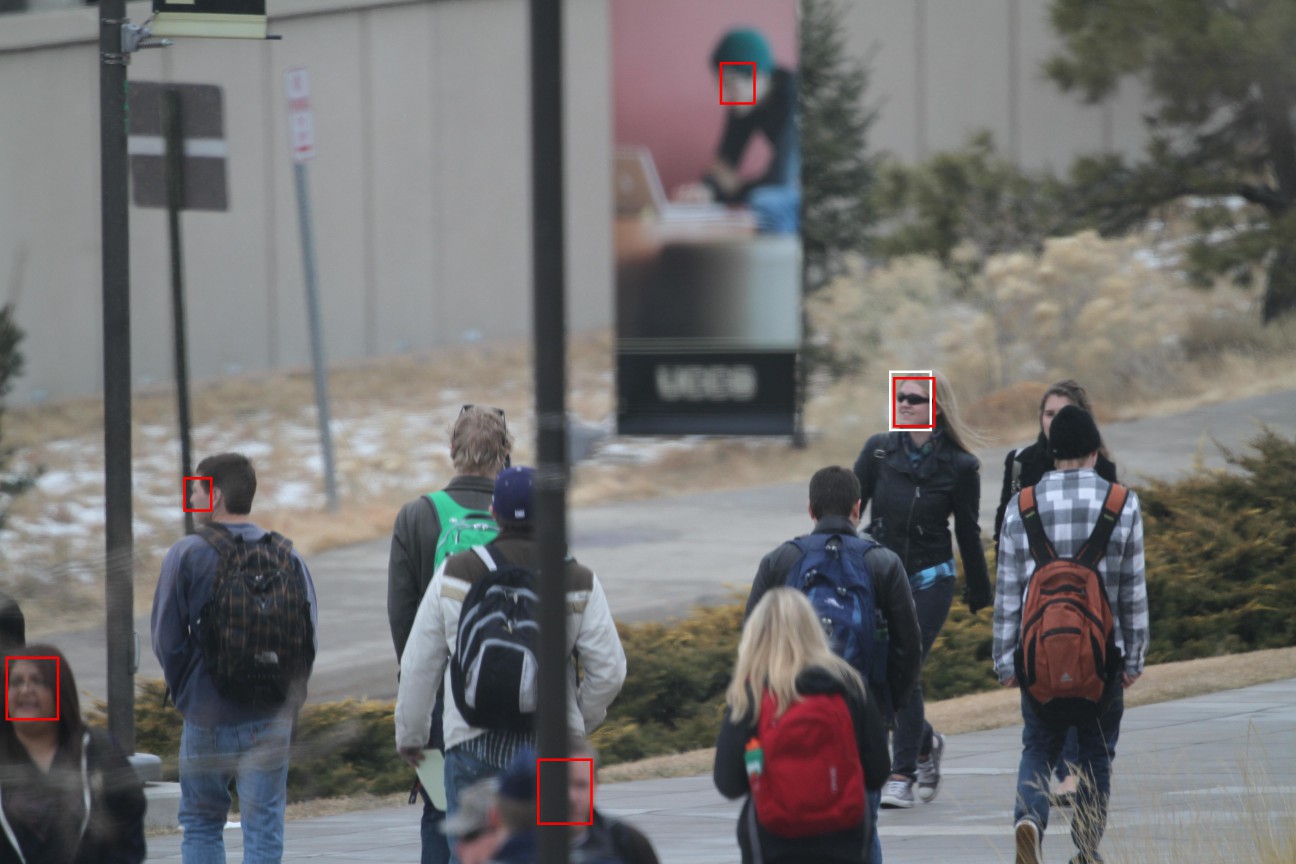} \\
\includegraphics[width=0.4\linewidth]{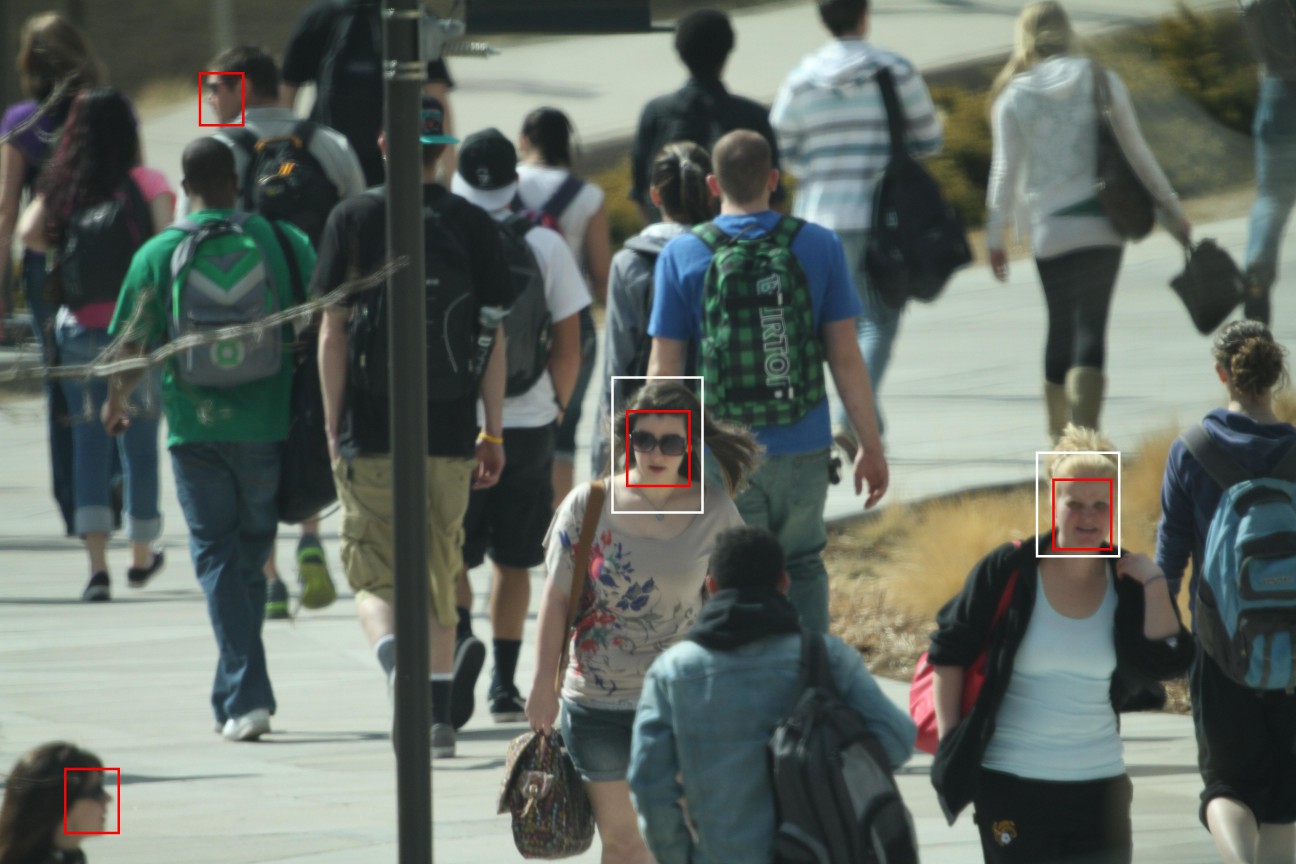} 
\includegraphics[width=0.4\linewidth]{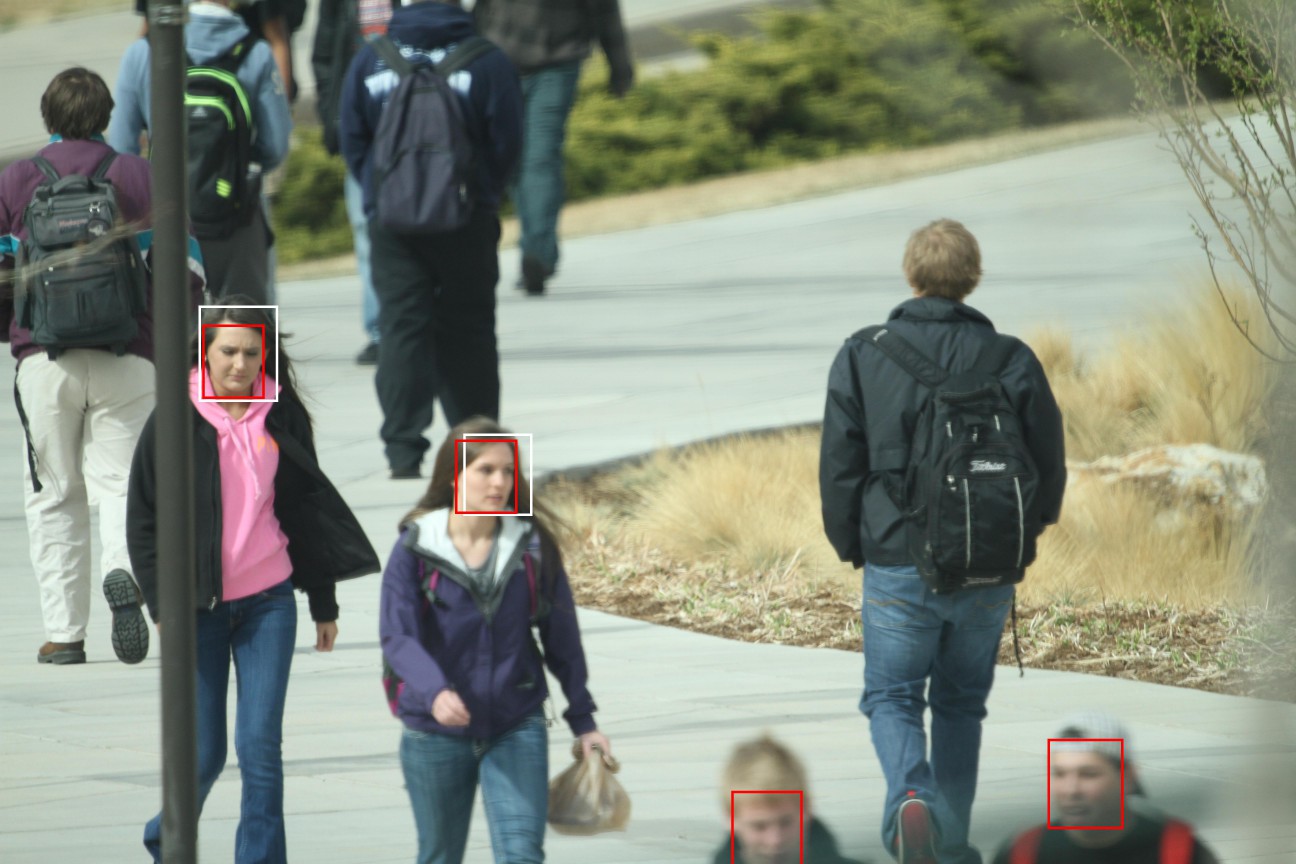}
\caption{Sample BFD based detection shown against ground truth. White bounding boxes are ground truth, and red ones are BFD detected faces. As you can see, BFD detect more faces than noted in ground truth. Specially those highly occluded and low quality faces.}
\label{fig:bfd_gt_samples}
\end{figure*}

\begin{figure}[!htbp]
\centering
\includegraphics[width=.5\linewidth]{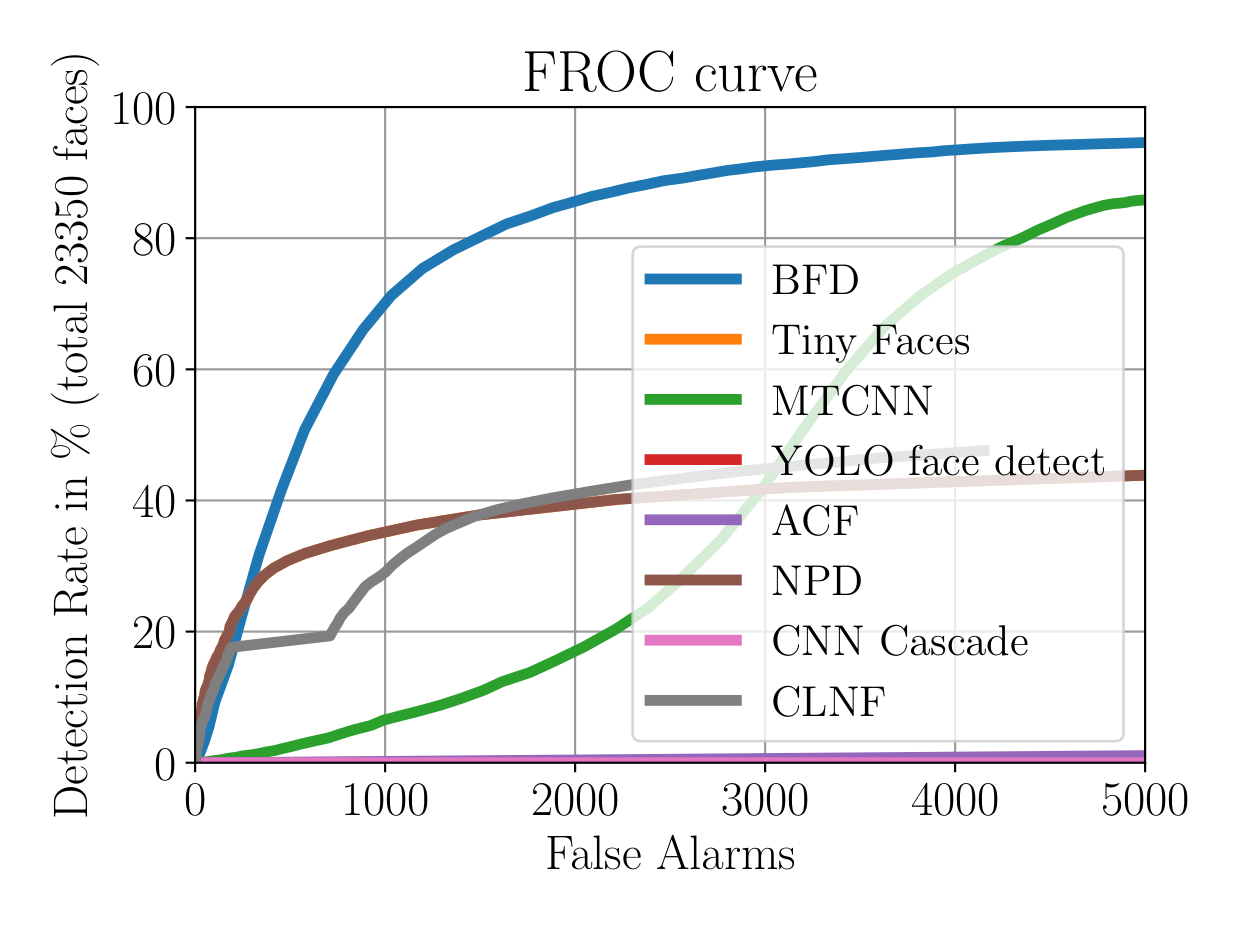}
\caption{Face Detection performance shown in FROC curve on training set.}
\label{fig:train_froc}
\end{figure}

\begin{figure}[!htbp]
\centering
\includegraphics[width=.5\linewidth]{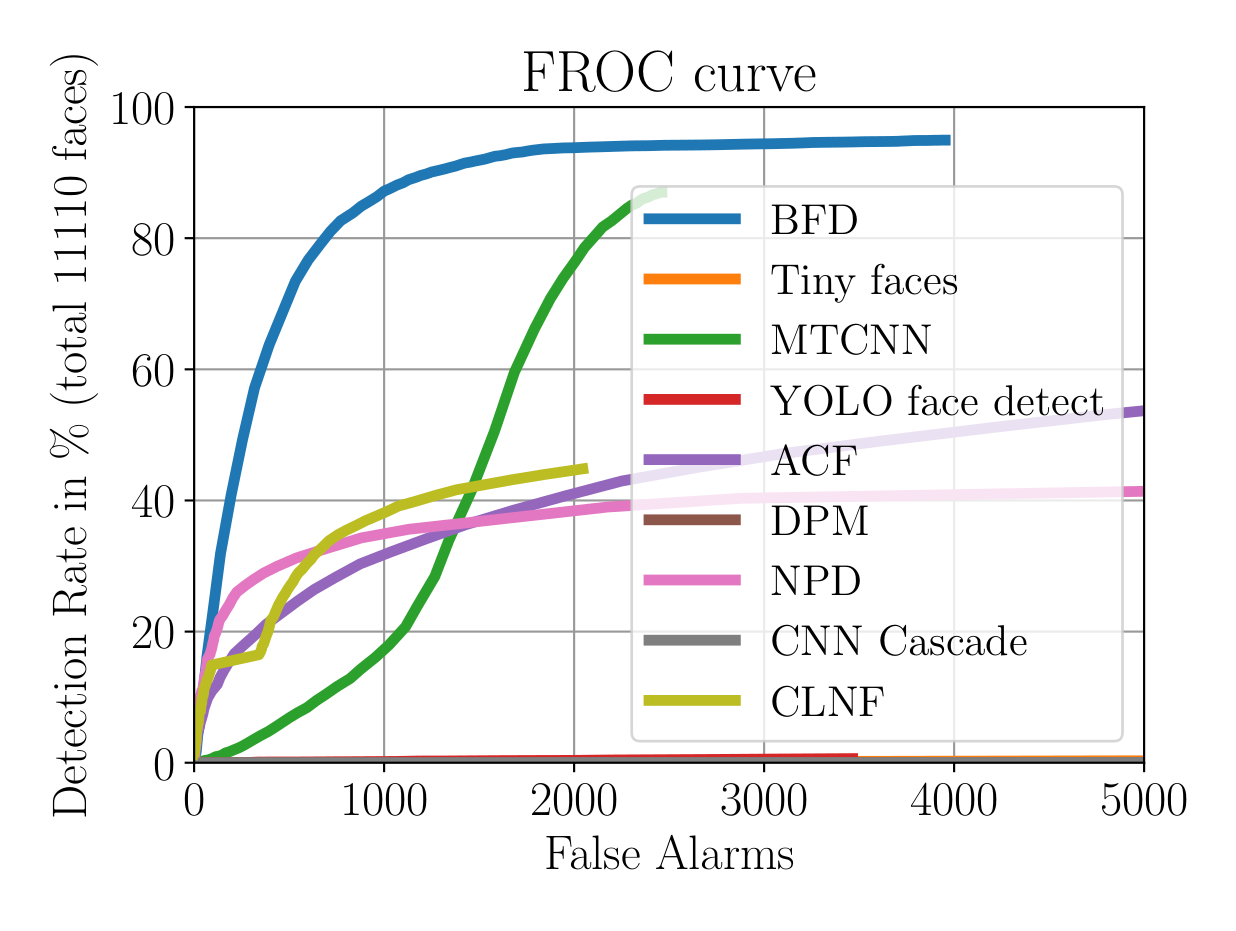}
\caption{Face Detection performance shown in FROC curve on validation set.}
\label{fig:valid_froc}
\end{figure}

\begin{figure}[!htbp]
\centering
\includegraphics[width=.5\linewidth]{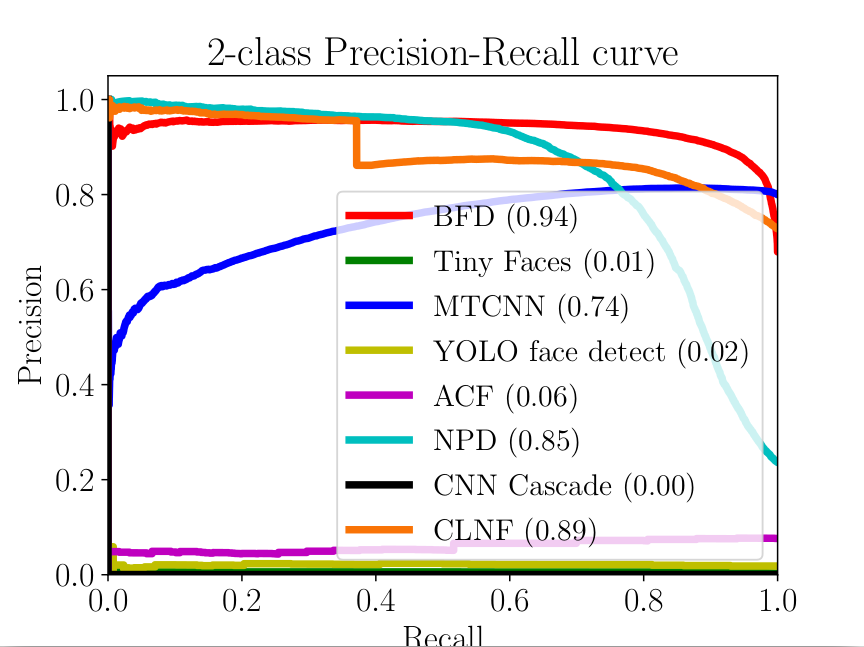}
\caption{Face Detection performance shown in Precision-Recall curve on training set. AUC is shown beside the method name within braces.}
\label{fig:pr_train}
\end{figure}

\begin{figure}[!htbp]
\centering
\includegraphics[width=.5\linewidth]{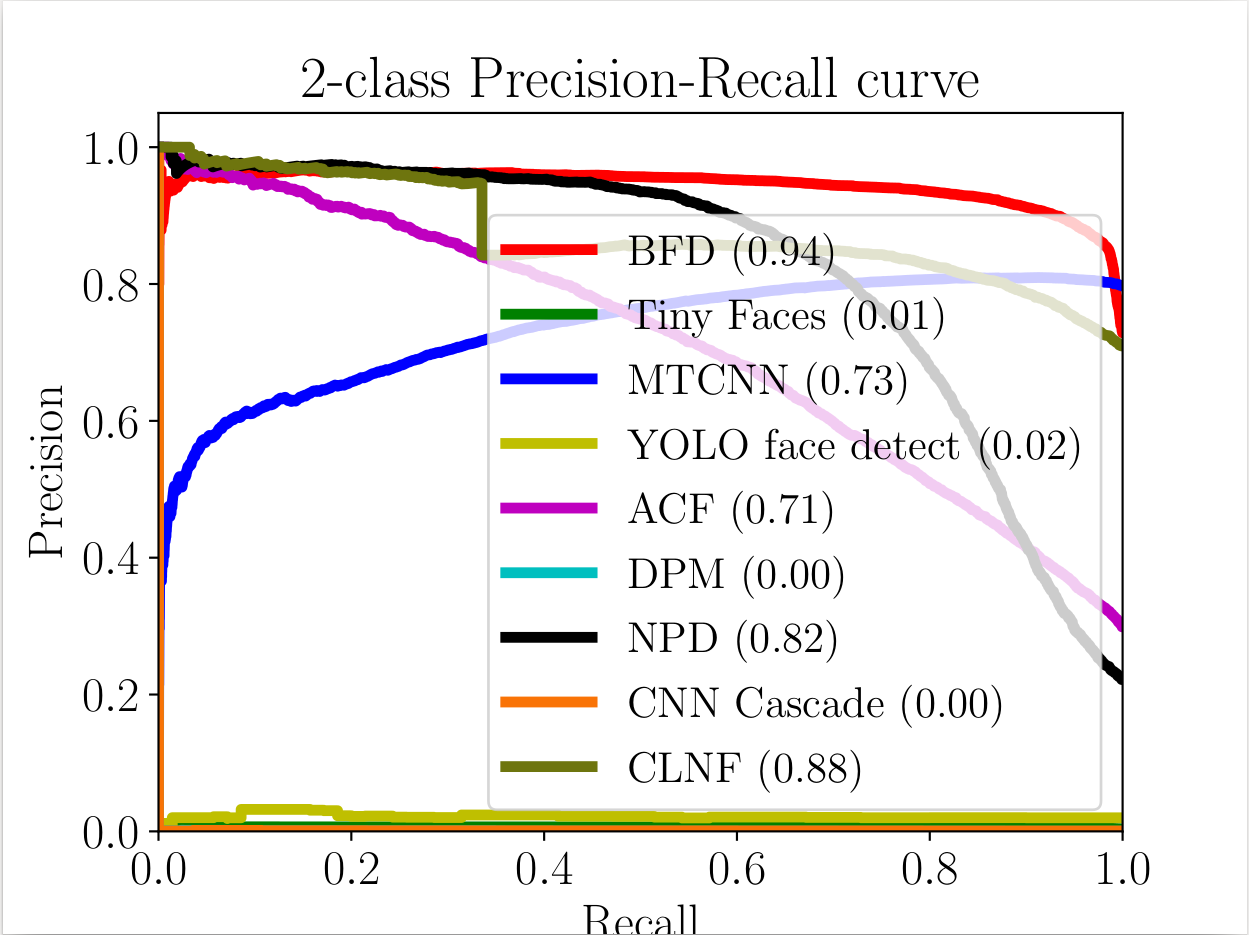}
\caption{Face Detection performance shown in Precision-Recall curve on validation set. AUC is shown beside the method name within braces.}
\label{fig:pr_valid}
\end{figure}

\begin{figure}[!htbp]
\centering
\includegraphics[width=.5\linewidth]{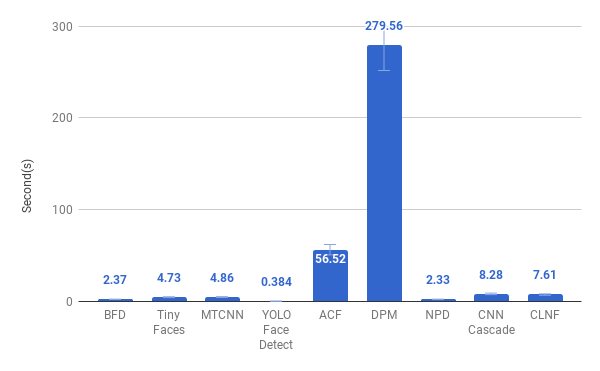}
\caption{Chart listing average face detection time per image}
\label{fig:time_chart}
\end{figure}

\section{Experiment Results}

For evaluation we use the training and validation set images of the UCCS database to perform face detection experiments using the previously mentioned approaches. The training set contains $10,917$ images with $23,350$ faces, and the validation set contains $5,237$ of $11,110$ faces. We don't use the test set because the database provides ground truth only for training and validation set. We use Free Receiver Operator Characteristic (FROC) to evaluate the face detection experiments and also provide detection details in Table~\ref{train_table} and Table~\ref{valid_table}. The face detection methods were used 'as-is' without any modification, if there was multiple options available we have tried to chose the optimal configuration available that produces the best accuracy.

Free Response Operating Characteristic (FROC) is based on a region-based analysis.  FROC analysis is similar to ROC analysis, except that the false positive rate on the $X$-axis is replaced by the number of false positives per image. Thus, FROC seeks location information from the observer (the algorithm), rewarding it when the reported face detection is marked in the appropriate location and penalizing it when it is not. Before FROC data can be analyzed, a definition of a detected region is needed. Although there are different opinions in the literature, the UCCS face detection challenge evaluation protocol use a typical approach which expects a $50\%$ overlap between the annotated and detected regions to indicate a true positive. 

In the FROC curves we can see the BFD method has the most AUC for false alarm with 5000 faces and significantly surpasses face detection methods. In Table~\ref{valid_table} it shows that Tiny Faces has the highest accuracy (0.966) but it also has 2.8M false alarms which makes it less effective for surveillance based detection scenario (the FROC curve is off the charts due to this reason). Our BFD method is the has 2nd highest accuracy (0.949) and 4th lowest false alarm ($3,953$) for validation set. The lowest false alarm is achieved by CLNF method but it has accuracy of $0.448$ which is very low. In Table~\ref{valid_table} we see BFD method has the highest accuracy ($0.959$) and 4th lowest false alarms ($10,887$). Both cases FROC curves shows best AUC for BFD method within 5000 false alarms. Lowest false alarm is achieved by also the CLNF method but again the accuracy is very low ($0.476$). In Figure~\ref{fig:time_chart} we show the average face detection time per image for each of the method discussed. We can see that YOLO Face Detect is the fastest , but we also know for both of the dataset it has performed very poorly. BF has the 3rd lowest detection time of $2.37$ seconds. Note that, we skip DPM method for Table~\ref{train_table} because it was very time consuming, we expect it to have similar accuracy as validation set.

In Figure~\ref{fig:pr_train} and Figure~\ref{fig:pr_valid} we provide the performance of the face detection methods for training and validation set in Precision-Recall curve. The precision-recall curve shows the trade-off between precision and recall for different threshold. The high area under the curve represents both high recall and high precision, where high precision relates to a low false positive rate, and high recall relates to a low false negative rate. We can see that, for Figure~\ref{fig:pr_train} BFD, NPD (0.85 in training and 0.82 in validation) and CLNF (0.89 in training and 0.88 validation set) shows the best AUC for both training set and validation set, with BFD showing the best performance of 0.94 in both training and validations set.

\begin{table}[!t]
\renewcommand{\arraystretch}{1.3}
\caption{Results on validation Set}
\label{valid_table}
\centering
\begin{tabular}{|c||c||c||c|}
\hline
Method & Detected & False Alarm & Accuracy\\
\hline
BFD & 10549 & 3953 & 0.949\\
\hline
Tiny Faces & 10740 & 876959 & 0.966 \\
\hline
MTCNN & 9671 & 2462 & 0.870 \\
\hline
YOLO Face Detect & 70 & 3466 & 0.006 \\
\hline
ACF & 7819 & 18341 & 0.703 \\
\hline
DPM & 32 & 134930 & 0.002 \\
\hline
NPD & 5373 & 18846 & 0.483 \\
\hline
CNN Cascade & 28 & 44104 & 0.002 \\
\hline
CLNF & 4980 & 2046 & 0.448 \\
\hline
\end{tabular}
\end{table}

\begin{table}[!t]
\renewcommand{\arraystretch}{1.3}
\caption{Results on training set}
\label{train_table}
\centering
\begin{tabular}{|c||c||c||c|}
\hline
Method & Detected & False Alarm & Accuracy\\
\hline
BFD & 22400 & 10887 & 0.959 \\
\hline
Tiny Faces & 12255 & 39622 & 0.524 \\
\hline
MTCNN & 20065 & 5028 & 0.859 \\
\hline
YOLO Face Detect & 128 & 6878 & 0.005 \\
\hline
ACF & 20834 & 252588 & 0.89 \\
\hline
DPM & ---  & --- & --- \\
\hline
NPD & 12255 & 39622 & 0.524 \\
\hline
CNN Cascade & 45 & 89128 & 0.001 \\
\hline
CLNF & 11115 & 4152 & 0.476 \\
\hline
\end{tabular}
\end{table}

We would like to further mention that, in Gunther et al. \cite{gunther2017uccs} $1000$ false alarms were detected on test set by both our body pose based method and Tiny Faces. We find it surprising because of the high number of false alarms generated by Tiny Faces in validation and training set. There is a possibility that they have tuned the algorithm further to get better accuracy for this database but due to lack of information provided we cannot say anything conclusively in this matter.

But why a body pose based method is more effective than generic face detection based methods for surveillance datasets ? In Figure~\ref{fig:bfd_gt_samples} we show some samples from the UCCS dataset to explain this. We know in survelliance scenario the camera is usually set at a distance to capture a wide angle and set above human height. This setting usually captures images with multiple faces with the whole body, because of the distance and the height of the position of the camera. For this reason we think, in this cases, taking the cue from the body detection can help us more to effectively find the face region. Because of surveillance scenario , the faces will be at a distance, prone to be affected by extreme pose angle, blurriness, high occlusion, varied lighting condition due to image taken at different times of the day. 

In Figure~\ref{fig:bfd_gt_samples} we have shown BFD detected faces in red bounding boxes and the ground truth in white. Notice that, several cases the ground truth misses to label highly occluded faces, and faces with extreme angles. But because of body based detection, our method was able to easily identify the face region.

\section{Conclusion}

Even though face detection research has come a long way and a lot of highly effective detection methods are already been published, we think there is still improvement needed in cases of surveillance scenario where most of the faces contain low biometric quality because of the nature of the camera settings (distance and height due to surveillance). In this work, we have compared several generic face detection methods and our proposed body based face detection method. Our results show the effectiveness of using body information for face detection for surveillance image databases.

\bibliographystyle{unsrt}  
%\bibliography{references}  %%% Remove comment to use the external .bib file (using bibtex).
%%% and comment out the ``thebibliography'' section.
\bibliography{IEEEabrv,uccs_face_detection}

%%% Comment out this section when you \bibliography{references} is enabled.
% \begin{thebibliography}{1}

% \bibitem{kour2014real}
% George Kour and Raid Saabne.
% \newblock Real-time segmentation of on-line handwritten arabic script.
% \newblock In {\em Frontiers in Handwriting Recognition (ICFHR), 2014 14th
%   International Conference on}, pages 417--422. IEEE, 2014.

% \bibitem{kour2014fast}
% George Kour and Raid Saabne.
% \newblock Fast classification of handwritten on-line arabic characters.
% \newblock In {\em Soft Computing and Pattern Recognition (SoCPaR), 2014 6th
%   International Conference of}, pages 312--318. IEEE, 2014.

% \bibitem{hadash2018estimate}
% Guy Hadash, Einat Kermany, Boaz Carmeli, Ofer Lavi, George Kour, and Alon
%   Jacovi.
% \newblock Estimate and replace: A novel approach to integrating deep neural
%   networks with existing applications.
% \newblock {\em arXiv preprint arXiv:1804.09028}, 2018.

% \end{thebibliography}

\end{document}